\documentclass{article}

\usepackage[nonatbib,preprint]{neurips_2021}

\usepackage[utf8]{inputenc} % allow utf-8 input
\usepackage[T1]{fontenc}    % use 8-bit T1 fonts
\usepackage{hyperref}       % hyperlinks
\usepackage{url}            % simple URL typesetting
\usepackage{booktabs}       % professional-quality tables
\usepackage{amsfonts}       % blackboard math symbols
\usepackage{nicefrac}       % compact symbols for 1/2, etc.
\usepackage{microtype}      % microtypography
\usepackage{xcolor}         % colors

\usepackage{graphicx}
\usepackage{epstopdf}
\usepackage{comment}
\usepackage{amsmath,amssymb} % define this before the line numbering.
\usepackage{color}

\usepackage{wrapfig}

\DeclareMathOperator*{\argmin}{arg\,min}

\title{Towards Photorealistic Colorization by Imagination}

\author{
Chenyang Lei$^*$ \qquad Yue Wu\thanks{Joint first authors} \qquad Qifeng Chen\\
The Hong Kong University of Science and Technology\\ 

}

\begin{document}

\maketitle

\begin{abstract}
We present a novel approach to automatic image colorization by imitating the imagination process of human experts. Our imagination module is designed to generate color images that are context-correlated with black-and-white photos. Given a black-and-white image, our imagination module firstly extracts the context information, which is then used to synthesize colorful and diverse images using a conditional image synthesis network (e.g., semantic image synthesis model). We then design a colorization module to colorize the black-and-white images with the guidance of imagination for photorealistic colorization. Experimental results show that our work produces more colorful and diverse results than state-of-the-art image colorization methods. Our source codes will be publicly available.
% Our network architecture leverages an off-the-shelf semantic segmentation to obtain semantic information and uses an segmentation-to-image network to imaging an example with similar semantic information of the input image. 
%Both colorization network and fusion module can be trained with large-scale data. 
\end{abstract}

\section{Introduction}
% What is the problem?
There exist numerous black-and-white photos captured decades ago, and there is a need for skilled artists to perform image colorization. However, such a manual process is labor-intensive and may take a long time to obtain photorealistic colorization results. In this work, we are interested in automatic image colorization that aims to produce a colored photorealistic image from a grayscale one by a computer program without any user guidance or reference. However, existing state-of-the-art methods have various weaknesses compared with human experts. The results by existing methods are often grayish or have color-bleeding artifacts, although many dedicated designs~\cite{Zhang2016,su2020instance} are proposed. Besides, most methods~\cite{Zhang2016,su2020instance,Larsson2016,Iizuka2016,Cheng2015} can only provide one colorization result.

%The performance of colorization is highly influenced by the types objects.

%The results are not co
%The results are not diverse. While the colorization results are more colorful with dedicated designs, we notice the results are grayish in some areas. 

% Why is it hard?
Why can human experts colorize black-and-white photos so well?\footnote{bit.ly/color\_history\_photos} This question is connected to diverse colors for the same kind of object. Intuitively, when we judge whether the color is reasonable, we have to understand the concept of objects. Similarly, when experts colorize a black-and-white photo, they understand what it is and imagine different colorful objects with the same concept. Also, experts can colorize black-and-white images by finding correspondences between the two objects (of the same concept). As a comparison, although direct colorization models are also designed to understand the contexts of an image, they only predict a color with the highest possibility for each pixel \cite{Zhang2016,Iizuka2016}. Hence, the automatic colorization performance varies a lot among different objects. When the color is closely related to the object category (e.g., green trees, blue sky), the performance is often satisfactory; for the other objects, the performance can be poor as there are multiple plausible colors, and the model often outputs the average color among all possible colors.
%since an average color (grayish color) corresponds to the lowest loss for all possible color.

\begin{figure}[t]
\centering
\includegraphics[width=1.0\linewidth]{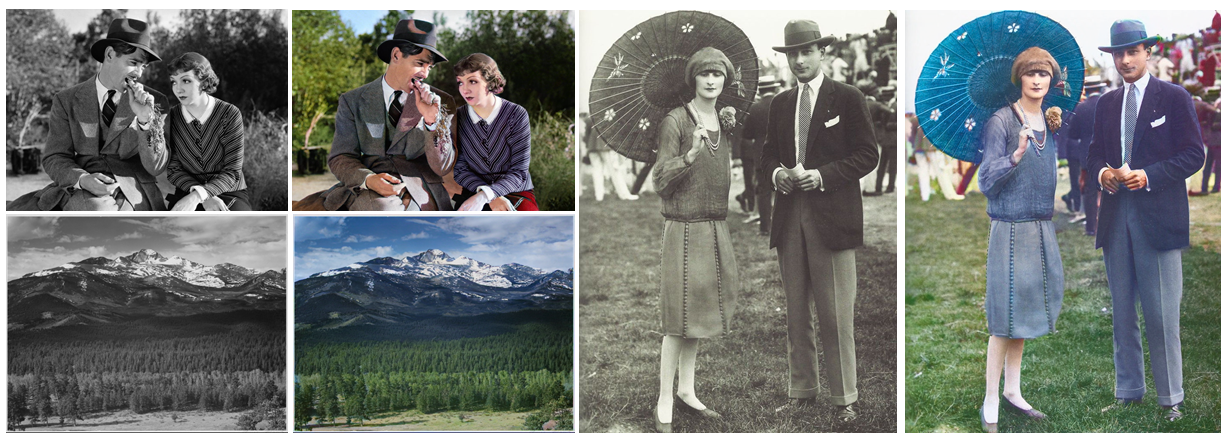}
\caption{Qualitative results on legacy black-and-white photographs. Our proposed approach can produce colorful and realistic colorized results. More results are presented in the supplementary material.}
\label{fig:legacy}
\end{figure}

% Why can our method work?
%and utilize the relationship between contexts and colors
By imitating the imagination of human experts, our approach can achieve photorealistic colorization for legacy black-and-white photos, as shown in Figure~\ref{fig:legacy} and Figure~\ref{fig:Teaser}.
% Our motivation is to imitate the imagination of human experts and utilize the relationship between contexts and colors for photorealistic colorization, as shown in Figure~\ref{fig:legacy}. 
We design an imagination module for understanding the contexts of images and imagining relevant visual content, similar to human imagination. The distribution of diverse and realistic colors can be embedded in the imagination module. Specifically, given a black-and-white image, we first predict the semantic layout or class of objects explicitly. Then we use an off-the-shelf generative model to synthesize images with the same semantic context, as shown in Figure~\ref{fig:Teaser}. The synthesized images are then used as the reference for colorization. Although the structures of synthesized images are not perfect, we notice that generative models can generate images with diverse and colorful colors.

%Note that the semantic of object with or without color should be the same in most cases. 

%Our approach is based on an assumption: there is an existing model that can generate diverse and colorful images. In practice, we notice this assumption can be achieved: although the structures of synthesized images are not perfect, their color is usually quite realistic.

% 
Instead of using a single image as the reference~\cite{He2018,zhang2019deepvideocolor}, we propose to compose a novel image through multiple images for better colorization results. This strategy is designed to solve a problem: even though the synthesized object and the input object share the same context/concept, the luminance of the synthesis object can be different from the input object. For example, given a black-red T-shirt, the synthesized T-shirt can be white. 

Extensive experimental results demonstrate that images colorized by our approach are more colorful and photorealistic than state-of-the-art methods. Besides, our results are quite diverse. We hope our imagination module can inspire tasks beyond colorization.
% On the one hand, synthesizing images with the semantic information lose some information: 

%Compared with single reference, using multiple references can reduce the possibility that there is no corresponding color objects in references images. 

%We believe this strategy can also be applied to the reference-based colorization methods that search the reference images from a big dataset: it is very possible that some objects in grayscale images do not appear in the single reference color image.

% We show that the colorized results can be quite diverse, colorful and realistic through the imagination module. An user study also demonstrates that our results are preferred by human observers compared with state-of-the-art methods.

% Extensive results show that results of our approach are diverse and colorful. 

\section{Related Work}

\textbf{User-guided colorization.}
Most classical approaches rely on user guidance for colorization, including scribble-based colorization and example-based colorization. Scribble-based colorization aims to propagate the color provided by users. Most methods~\cite{Levin2004,Qu2006,Luan2007,Daniel2009lazy,Chen2012,Yatziv2006,Patsorn2017Scribbler,lvmin2018Two,Filling2021zhanguser} achieve this goal based on optimization with some assumption. For example, Levin et al.~\cite{Levin2004} assume pixels with similar intensity should have similar colors and design an optimization approach based on this objective. In addition to optimization based approach, Zhang et al. \cite{Zhang2017} propose a learning-based model for scribble-based colorization. Compared with previous work, fewer scribbles are required in their method.

% using user input on part of the image to propagate the provided colors on certain regions to the whole image \cite{Levin2004,Qu2006,Luan2007,Chen2012,Yatziv2006}. Levin et al. \cite{Levin2004} propose optimization based interactive image colorization by solving a quadratic cost function under the assumption that similar pixels in space-time should have similar colors. 

Example-based colorization~\cite{Welsh2002,Ironi2005,Liu2008,Charpiat2008,Chia2011,Gupta2012,He2018,zhang2019deepvideocolor} further reduces the difficulty of colorizing a black-and-white photo for normal users. Only an additional reference image is required to colorize images automatically. Since the reference image is usually closely related to grayscale images, these methods can find the correspondence between two images. However, it is non-trivial to find a good reference image since the reference image should be semantically similar to the input grayscale image. User needs to search the image on the Internet or provide semantic text labels and segmentation cues to an image retrieval system~\cite{Chia2011}, which is not user-friendly. He et al. \cite{He2018} introduce deep learning to example-based colorization and adopt an image retrieval algorithm to search an image from a large dataset. While the performance of colorization is improved a lot, searching for a good reference is still time-consuming for them. In this work, our imagination module avoids the efforts to find a reference image.

\textbf{Automatic colorization.} Deep learning methods~\cite{Cheng2015,Deshpande2015,Iizuka2016,Zhang2016,Larsson2016,Deshpande2017diverse,safa18Structural,Seungjoo2019Limited,Lei_2019_CVPR,Patricia2020ChromaGAN,Jiaojiao2020Pixelated,Manoj2021transformer,su2020instance} achieve fully automatic colorization by learning the relationship between semantic and color on a large dataset. For automatic colorization, the performance is usually good when the color of the object (e.g., a green tree) is closely related to the semantic~\cite{Zhang2016,Larsson2016}. However, for the other objects that can have diverse colors (e.g., T-shirt), the colorization results are usually grayish. While treating the colorization as a classification problem~\cite{Zhang2016,Zhang2017,su2020instance} can improve the colorfulness, these cases are still not rare. Compared with them, our model can generate diverse and colorful results with the help of our imagination module.

\textbf{Conditional image generation.}
The generative models are classified into two categories: unconditional and conditional. From the aspect of unconditional generative models, the most recent work is StyleGAN~\cite{stylegan} and StyleGAN2~\cite{stylegan2}, which are based on style transfer literature and achieve state-of-the-art performance for high-resolution image synthesis. From the aspect of conditional image synthesis, BigGAN~\cite{biggan} use object class and a random noise as input, and can synthesis high-quality images. Another source of conditional GANs takes more concrete conditions, such as layout~\cite{DBLP:conf/iccv/SunW19,DBLP:journals/corr/abs-2003-11571}, semantic map~\cite{DBLP:conf/iccv/LiZM19,wang2018pix2pixHD,park2019semantic,oasis} et al. These generative models capture the distribution of a dataset and can be utilized to help solve colorization problem.

% The most prominent work on fully automatic image colorization is deep learning based approaches that do not require any user guidance \cite{Cheng2015,Iizuka2016,Zhang2016,Larsson2016,Deshpande2017}. Cheng et al. \cite{Cheng2015} propose the first deep neural network model for fully automatic image colorization. Some deep learning approaches use a classification network that classifies each pixel into a set of hundreds of chrominance samples in a LAB or HSV color space to tackle to the multi-modal nature of the colorization problem \cite{Zhang2016,Larsson2016}. However, it is difficult to sample densely in the two-dimensional chrominance with hundreds of points. Thus we propose to use a perceptual loss with diversity \cite{Li2018} to avoid the discretization problem.

\section{Overview}
Given a grayscale image $\mathbf{X} \in \mathbb{R}^{H \times W \times 1} $ as input, our goal is to automatically predict its color channels $\mathbf{Y} \in \mathbb{R}^{H \times W \times 2} $. $\mathbf{X}$ and $\mathbf{Y}$ are the lightness and $ab$ channel in the CIE $Lab$ color space. The key difference between our approach and existing automatic colorization methods is that: we propose to synthesize color images based on the context of the input image as a reference for colorization.

\begin{figure}[t]
\centering
\begin{tabular}{@{}c@{}}
\includegraphics[width=1.0\linewidth]{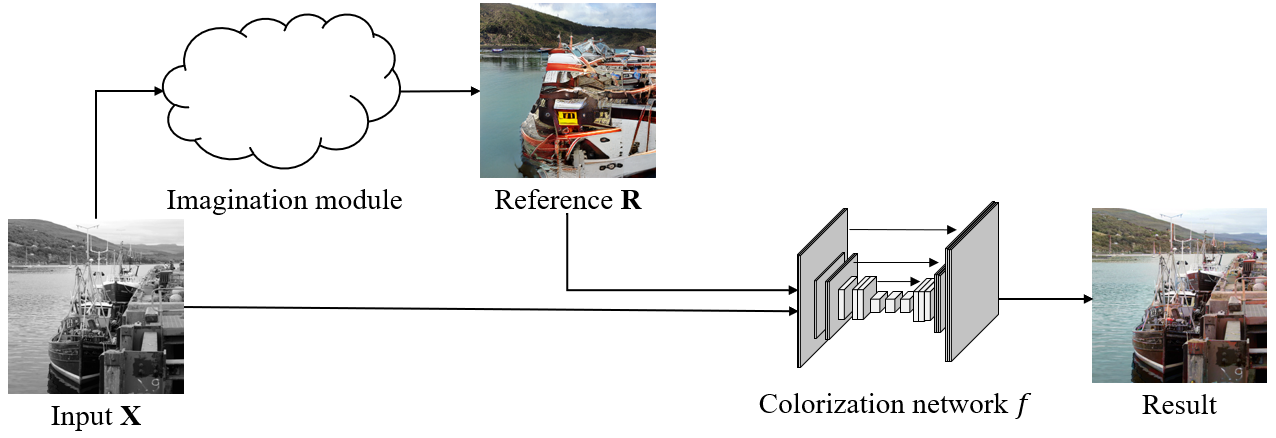}\\
% Lai et al.~\cite{lai2018learning}\\
\end{tabular}
\caption{The overall framework of our approach. In our imagination module, we firstly synthesize a reference image $\mathbf R$ that has the similar context with the input grayscale image $\mathbf X$. In our colorization network, $\mathbf X$ is colorized with the guidance of reference image $\mathbf R$. } 
\label{fig:overview}
\end{figure}

Figure~\ref{fig:overview} shows the architecture of our approach. From the imagination module, we obtain a colorful reference image $\mathbf{R} \in \mathbb{R}^{H \times W \times 3} $ with coarsely aligned context information for a grayscale image $\mathbf{X}$. In our imagination module, we first extract the context and location information through off-the-shelf methods (e.g., semantic segmentation, detection, classification). Then, we can synthesize images based on the extracted information using a generative model. The details of the imagination module is described in Section 4.

As shown in Figure~\ref{fig:overview}, the grayscale image $\mathbf{X}$ and synthesized reference image $\mathbf{R}$ are then used as input to our colorization network $f$ to predict the color channels $\mathbf{\hat Y}$. The colorization network $f$ aligns reference color and grayscale images. It keeps the original correct color and removes undesirable artifacts simultaneously. The training strategy and details of network $f$ are presented in Section 5.

\begin{figure*}[t]
\centering
\begin{tabular}{@{}c@{\hspace{1mm}}c@{\hspace{1mm}}c@{\hspace{1mm}}c@{\hspace{1mm}}c@{\hspace{1mm}}c@{\hspace{1mm}}c@{}}
&
\includegraphics[width=0.191\linewidth]{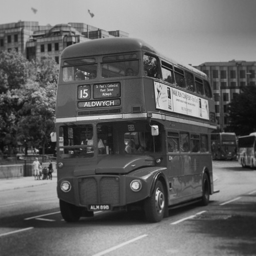}&
\includegraphics[width=0.191\linewidth]{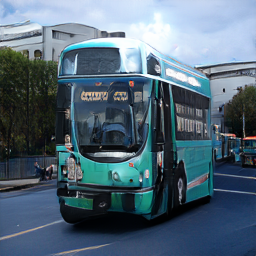}&
\includegraphics[width=0.191\linewidth]{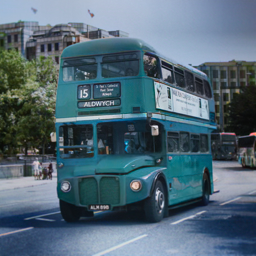}&
\includegraphics[width=0.191\linewidth]{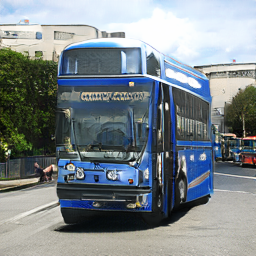}&
\includegraphics[width=0.191\linewidth]{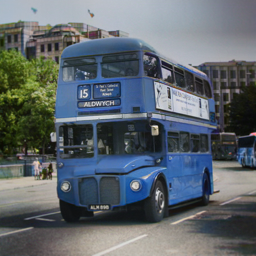}\\
&
\includegraphics[width=0.191\linewidth]{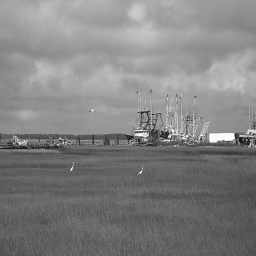}&
\includegraphics[width=0.191\linewidth]{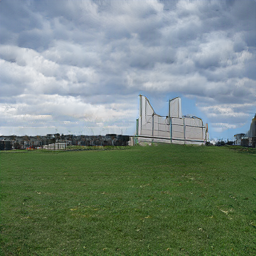}&
\includegraphics[width=0.191\linewidth]{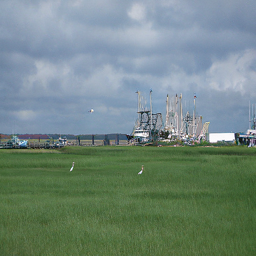}&
\includegraphics[width=0.191\linewidth]{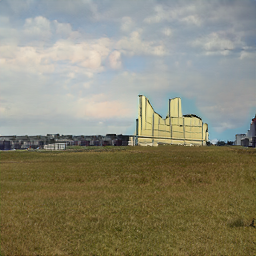}&
\includegraphics[width=0.191\linewidth]{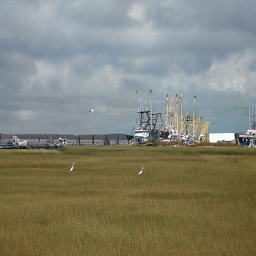}\\
 &Input & Imagination 1 & Result 1  & Imagination 2 & Result 2  \\ 

\end{tabular}
\caption{Example grayscale images and colorized results of our approach. Our proposed method colorizes gray images by imaging color images with similar contexts for automatic colorization. Our colorized results are diverse and colorful.}
\label{fig:Teaser}
\end{figure*}

\section{Imagination Module}
\subsection{Objective}
Given a grayscale image $\mathbf{X}$, our imagination module aims to generate images $\mathbf{R}$ as references for photorealistic colorization. The motivation is to imitate human imagination: experts can produce photorealistic colorized results since they know diverse and colorful examples for the black-and-white objects. Hence, the images $\mathbf{R}$ generated by our imagination module must be diverse, colorful, and realistic since the color of the final colorized image is mainly decided by our imagination $\mathbf R$.

Specifically, we need to understand the context of the grayscale image $\mathbf{X}$ first. Then we generate images $\mathbf{R}$ using the extracted context $\mathbf{S}$, as described in Section 4.2. Take Figure~\ref{fig:Teaser} as an example; our model first understands that there is a car in the grayscale image $\mathbf{X}$. Then, our model can `imagine' cars with diverse colors. Since the color of synthesized images is closely related to the contexts, the color is usually reasonable, and we can colorize the grayscale images with guidance.

%Our key observation is that image synthesized by GAN without structure-regularization is colorful and diverse. By taking the semantic information of grayscale image $\mathbf{X}$ into consideration, the synthesized image can provide reliable guidance for colorization.

\textbf{Comparison to image retrieval.} He et al.~\cite{He2018} propose a novel image retrieval algorithm that automatically recommends good references to the user. Compared with them, our approach is more efficient in memory and time. Firstly, we do not need a large-scale dataset (e.g., ImageNet has more than 1 million images) in the inference time. Secondly, we do not need to spend time searching for the most similar reference. 

\textbf{Comparison to GAN inversion.} Another possible solution for synthesizing a reference image is GAN-inversion, as demonstrated by Pan et al.~\cite{pan2020dgp}. However, we notice that images obtained by GAN inversion are not satisfying: the contents are wrong in many cases (and hence, colors are wrong). We believe it is because only structure information is utilized in the optimization process, and the degradation operation (color to gray) makes this problem even more difficult. Some examples can be found in the supplementary material.

\subsection{Context extraction and image generation}
Given a grayscale image $\mathbf X$, the pipeline to imagine a reference image $\mathbf R$ can be describe as follows:
\begin{align}
    \mathbf R = g(c(\mathbf X), \mathbf{z}),
\end{align}
where $c$ is the context extraction model, $g$ is the image generation model, and $\mathbf{z}$ is the latent code for the image generation model $g$. 

\textbf{Representation of context.} The representation of context can be quite diverse in computer vision. The key is that the model needs to know what the object is. In addition, it would be better if there is an existing image synthesis model $g$ based on that representation. Generating an image with context can be achieved by various methods, including semantic image synthesis~\cite{park2019semantic,oasis}, layout to image generation~\cite{zhao2019image}, text to image synthesis~\cite{reed2016text}, etc. In this paper, we mainly choose semantic segmentation as the representation of context. 

\textbf{Implementation.} 
Since we use the semantic segmentation to represent the context, the context extraction model $c$ is a semantic segmentation method, and the image generation model $g$ is the semantic image synthesis method.

\begin{figure}[t]
\centering
\begin{tabular}{@{}c@{}}
\includegraphics[width=1.0\linewidth]{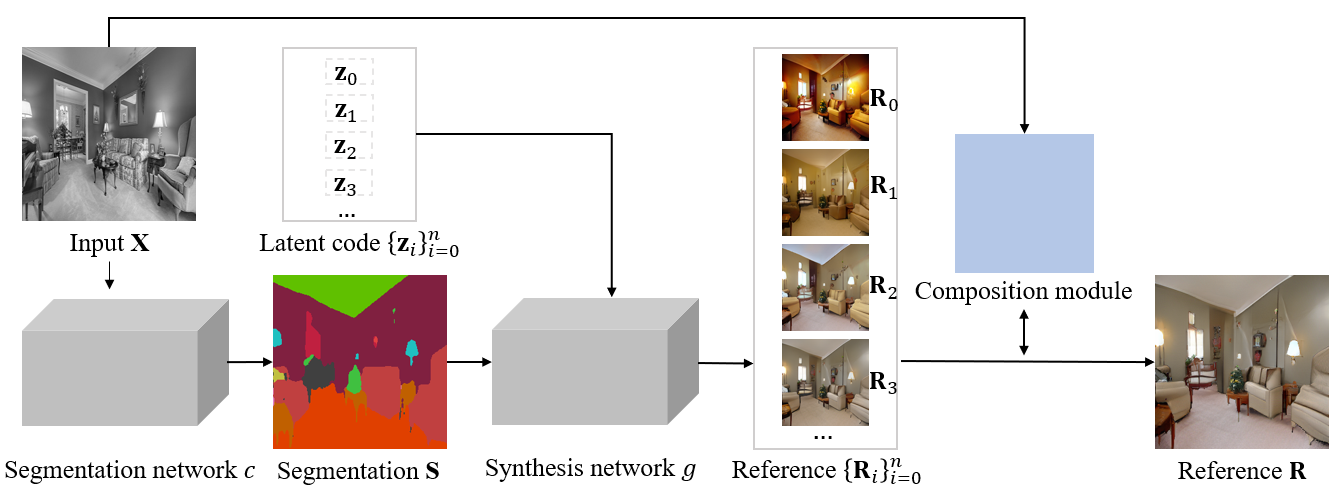}\\
% Lai et al.~\cite{lai2018learning}\\
\end{tabular}
\caption{The imagination module of our approach. First, we extract the segmentation of the grayscale image $\mathbf{X}$ with a segmentation network $c$. Secondly, we synthesize diverse images $\{\mathbf{R}_i\}_{i=0}^n$ by using the semantic segmentation $\mathbf{S}$ and latent codes $\{\mathbf{z}_i\}_{i=0}^n$. At last, our composition module composes a reference $\mathbf{R}$ from all images $\{\mathbf{R}_i\}_{i=0}^n$ for each segmentation based on the luminance of input $\mathbf{X}$. }
\label{fig:imagination}
\end{figure}

As shown in Figure~\ref{fig:imagination}, given a grayscale image $\mathbf{X}$, we implement semantic segmentation to obtain the segmentation $\mathbf{S}$ using an off-the-shelf model by $\mathbf{S}=c(\mathbf{X})$. The number of segments is different in each image, and we denote each segmentation as $\mathbf{S}^j$. Given the segmentation $\mathbf{S}$, we can synthesize a reference image $\mathbf{R}_i$ through a semantic image synthesis model $g$ by:
\begin{align}
\mathbf{R}_i=g(\mathbf S, \mathbf{z}_i),    
\end{align}
where the diversity of reference image $\mathbf{R}_i$ is controlled by the latent code $\mathbf{z}_i$. Note that we can sample an arbitrary number of reference images $\mathbf{R}_i$ using various $\mathbf{z}_i$. We denote $\mathbf{R} _i^j$ as the area of the reference image $\mathbf{R}_i$ in each segmentation $\mathbf{S}^j$. 

%In this paper, we demonstrate our idea mainly using the semantic information since we hope the objects between grayscale image $\mathbf{X}$ and reference image $\mathbf R$ are coarsely aligned. In the experiments (Section 6), we also present examples using detection, classification and class to image synthesis to achieve our goal.

% Synthesizing a good reference image requires understanding the semantic information of the grayscale image $\mathbf{X}$. Given a grayscale image $\mathbf{X}$, we first predict the semantic segmentation $\mathbf{S}$. As an off-the-shelf image generation module captures the color distribution of natural images conditioned on semantic classes, the image generation module is adopted to generate the reference image $\mathbf R$. $\mathbf R$ has a similar architecture with $\mathbf{X}$ and provides reasonable and colorful color.

\textbf{Diversity of imaginations}. While mode collapse~\cite{srivastava2017veegan} is still an open problem, we observe that image synthesized by generative models without structure-regularization is colorful and diverse. As shown in Figure~\ref{fig:Teaser}, although the structures of objects look poor, the color is relatively realistic and colorful.

\subsection{Composition module}
Although a single reference image $\mathbf R_i$ is is enough in many cases, we improve the performance further using multiple references to solve a problem: while $\mathbf R_i$ has a similar architecture with $\mathbf X$, there might be some incompatible segmentations where $\mathbf R^j_i$ mismatches with $\mathbf{X}^j$ in terms of luminance. For example, a desk within high luminance tends to be bright color, while the image generation model may provide a dark color desk. 

We address this problem by composing a new image through sampling multiple reference images $\{\mathbf{R}_i\}_{i=0}^n$, where $n$ is the number of samples images. For each segmentation $\mathbf{S}^j$, we aim to find the best reference segmentation from all candidates $\{\mathbf{R}_i^j\}_{i=0}^n$. Specifically, for each segmentation $\mathbf{S}^j$, we select the nearest neighbor among all references in terms of the average difference in luminance:
\begin{equation}
    \beta(j) = \argmin_i \sum_p {\left \| \mathbf R_{i}^j(p)-\mathbf X^j(p) \right \|_1 },
    % j = argmin \frac{\left \| R_{j}-X_i \right \|_1 }{\left \| m(S^j) \right \|_1 }
\end{equation}
where $p$ is the pixel and $\beta(j)$ denotes the best index for segmentation $\mathbf{S}^j$. At last, a final reference image $\mathbf{R}$ can be obtained through the set $\{\mathbf R_{\beta(j)}^j\}_j$.

\textbf{Interactive composition.} So far, all the steps can be implemented automatically. Here we describe how to allow interactive colorization. When the user is unsatisfied about an object $\mathbf X^j$ in the image, we can directly remove $\mathbf R_{\beta(j)}^j$ from the current $\{\mathbf R_{\beta(j)}^j\}_j$. We can also replace a segmentation with another one. For example, in Figure~\ref{fig:Teaser}, we can replace the blue car with the green car and preserve the colors in the other area.

% \begin{equation}
%     R = \sum_{p} m(p) \cdot \mathbf{R}_i
% \end{equation}
% where $m(p)$ is a mask indicate which pixels belongs to class $p$.

\section{Colorization by Imagination}
Our colorization module takes the original grayscale image $\mathbf X$ and a reference image $\mathbf R$ as input. The overall goal of this module is similar to reference-based colorization: we need to extract the color from the reference image $\mathbf R$. However, our reference $\mathbf R$ is different from natural images. Our $\mathbf R$ has the advantage that objects are coarsely aligned, but the detailed structures of synthesized images are inconsistent with grayscale images $\mathbf X$. Some artifacts might also appear in the reference image $\mathbf R$. Briefly speaking, it is impossible to have a perfect reference image since models in the imagination modules cannot be perfect (e.g., semantic segmentation and semantic image synthesis). Hence, we design a strategy to train a dedicated network that produces photorealistic results.
%Our colorization network is designed to copy the color from the reference frame based on the similarity between $\mathbf{X}$ and $\mathbf R$. Also, it should be able to ignore the unreasonable guidance. We directly adopt the perceptual loss~\cite{johnson2016perceptual} to optimize the network $f$.
% \subsection{Color alignment module}
% As the internal structure of a semantic segment is diverse and uncertain, we adopt a alignment module~\cite{zhang2019deepvideocolor} to align the reference and grayscale image. Firstly, we extract the features of $\mathbf{X}$ and $\mathbf R$, denoted as $F(X), F(R)$. For each semantic segments $\mathbf{S}^j$, where $j$ denotes the index of semantic class, we calculate the dense correspondence by computing the pairwise similarity between $F_X$ and $F_R$. $F_X$ and $F_R$ are normalized.

% The correlation matrix $M \in \mathbb{R}^{HW \times HW}$ measures the similarity between $F_X$ at position $i$ and $F_R$ at position $j$. The similarity of pixels not belonging to this semantic class is denotes as -inf as invalid.
% The we warp the reference color $R_{ab}$ towards $\mathbf{X}$ according to the correlation matrix. 
% \begin{align}
%         W^{ab}(i) = \sum_{p} m(p)\cdot \sum_{j} \mathop{softmax(C(i,j))}\limitS^j \cdot R_{ab}(j)
% \end{align}
% where $m(p)$ indicate which pixel belong to semantic class $p$. This process approximates selecting the pixel in the reference image within largest similarity score and belonging to the same class. The $W^{ab}$ serves as an aligned color reference to guide the colorization in next procedure.

\subsection{Training}

We need to generate simulated reference images $\mathbf R'$ that mimic artifacts of synthesized images $\mathbf R$ at test time. The details for simulating $\mathbf{R'}$ is introduced at the end of this section. In the training time, we predict color channels $\mathbf{\hat Y}$ directly given the grayscale image $\mathbf{X}$ and simulated reference $\mathbf R'$:
\begin{align}
    \mathbf{\hat  Y} = f(\mathbf X,w(\mathbf R');\theta_f),
\end{align}
where $\theta_f$ is the parameters of the colorization network, $w$ is the warp function to align $\mathbf{R}'$ and $\mathbf X$, following Zhang et al.~\cite{zhang2019deepvideocolor}. The network $f$ is trained with a perceptual loss:
\begin{align}
    L(\theta_f) =\sum_l \lambda_l{\|\Phi_l( r(\mathbf X,  \mathbf{\hat Y})) - \Phi_l(r(\mathbf{X}, \mathbf Y))\|_1},
\end{align}
where $\Phi_l$ is the feature maps for $l$-th layer of a pretrained VGG-19 network~\cite{simonyan2014very} and $\lambda_l$ is the weights for each layer. We adopt 5 layers 'conv1\_2', 'conv2\_2', 'conv3\_2', 'conv4\_2', and 'conv5\_2' in practice. $r$ is the function to convert image from Lab space to RGB space. We adopt the U-Net~\cite{Ronneberger2015} as the architecture of the colorization network $f$.

Our training strategy is different from previous reference-based colorization methods~\cite{He2018,zhang2019deepvideocolor}. The comparison is presented in the experiments.

\textbf{Data simulation.}
Our key idea is to ensure the simulated image $\mathbf R'$ has the similar color with ground truth. Otherwise, while the reference image $\mathbf R$ has reasonable color, it might be different from the ground truth. In this case, we observe the training process is unstable and results tend to reduce the colorfulness like previous work~\cite{He2018}.

Given a grayscale image $\mathbf X$ and its corresponding color channels $\mathbf Y$, we randomly sample the location and area to obtain a mask $\mathbf M \in \{0,1\}^{H \times W \times 1}$. Then we synthesize a new image $\mathbf Y'$:
\begin{align}
    \mathbf Y' = \mathbf Y \odot (1 -\mathbf  M) + \mathbf Y^{fake} \odot \mathbf M,
\end{align}
where $\mathbf Y^{fake}$ is the color channels from a irrelevant image. The simulated reference image $\mathbf R'$ is then obtained by $\mathbf R' = r(\mathbf X, \mathbf Y')$. Our motivation is to preserve the original color of reference $\mathbf R$ if the color is reasonable and remove the artifacts if the color is wrong.

\begin{table}[t!]
\centering
\setlength{\tabcolsep}{3mm}
\begin{tabular}{@{}l@{\hspace{4mm}}c@{\hspace{2mm}}c@{\hspace{2mm}}c@{\hspace{2mm}}c@{\hspace{2mm}}c@{\hspace{2mm}}c@{\hspace{2mm}}c@{\hspace{2mm}}c@{}}
\toprule
Comparison & \small{CIC~\cite{Zhang2016}}&\small{Iizuka~\cite{Iizuka2016}} & \small{Larrson~\cite{Larsson2016}}& ~\small{Deoldify~\cite{deoldify}}&\small{Lei~\cite{Lei_2019_CVPR}}&\small{Zhang~\cite{Zhang2017}} &\small{Su~\cite{su2020instance}} &Ours \\
\midrule
Colorfulness $\uparrow$  &0.158 &0.116&0.079&0.087&0.057&0.114&0.116& \textbf{0.184} \\
Diversity  & $\times$ &$\times$&$\times$&$\times$&\checkmark& $\times$& $\times$&\checkmark \\
\bottomrule
\end{tabular}
\vspace{2mm}
\caption{Comparisons to existing automatic colorization methods. }
\label{table:baseline_comparison}
\end{table}

\begin{figure*}[t]%[p]
\centering
\begin{tabular}{@{}c@{\hspace{1mm}}c@{\hspace{1mm}}c@{\hspace{1mm}}c@{\hspace{1mm}}c@{\hspace{1mm}}c@{}}

&\includegraphics[width=0.19\linewidth]{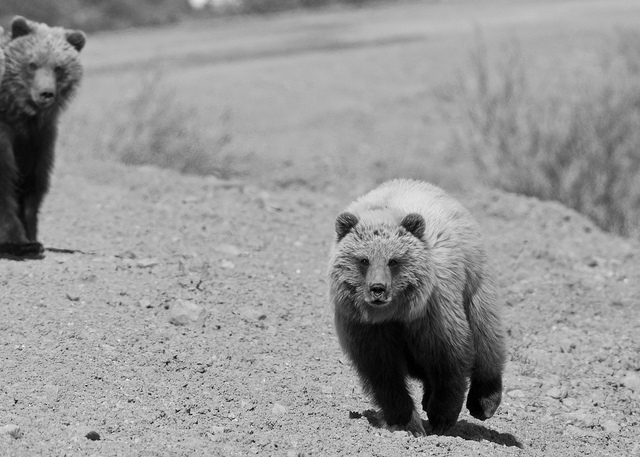}&
\includegraphics[width=0.19\linewidth]{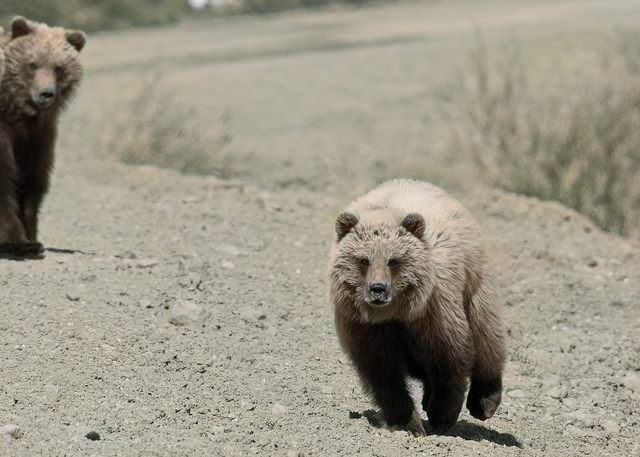}&
\includegraphics[width=0.19\linewidth]{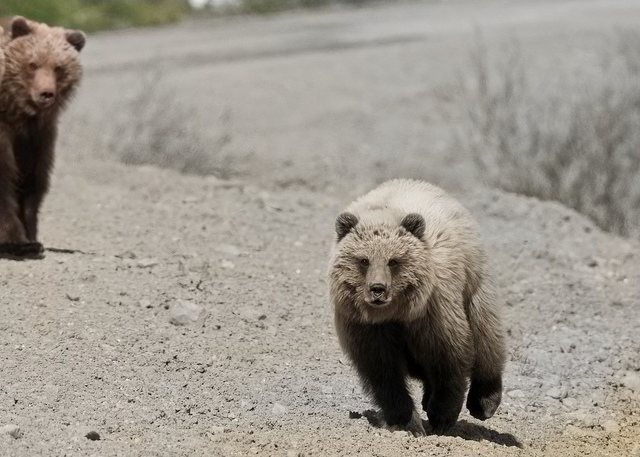}&
\includegraphics[width=0.19\linewidth]{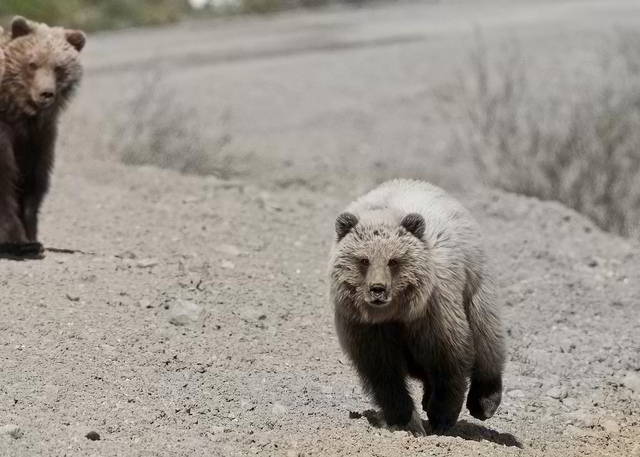}&
\includegraphics[width=0.19\linewidth]{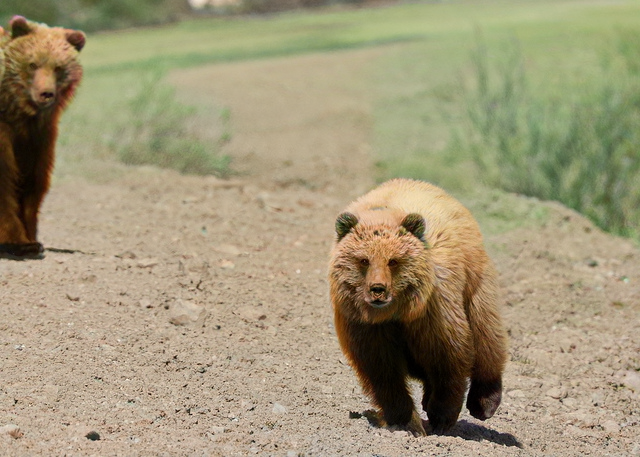}\\

&\includegraphics[width=0.19\linewidth,height=0.16\linewidth]{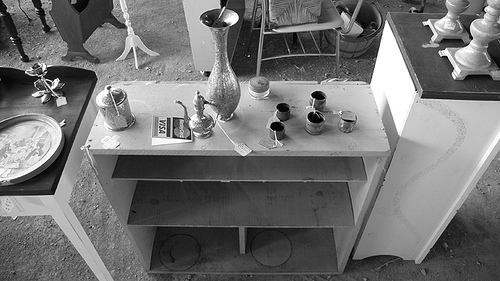}&
\includegraphics[width=0.19\linewidth,height=0.16\linewidth]{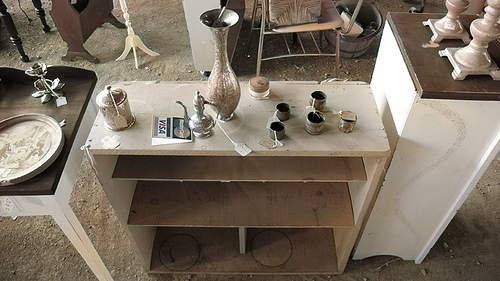}&
\includegraphics[width=0.19\linewidth,height=0.16\linewidth]{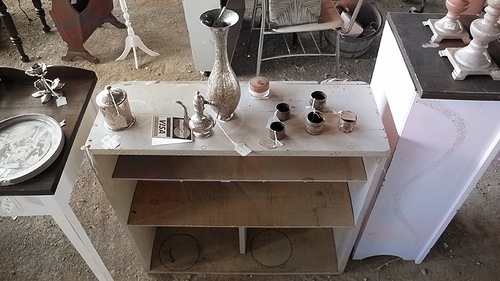}&
\includegraphics[width=0.19\linewidth,height=0.16\linewidth]{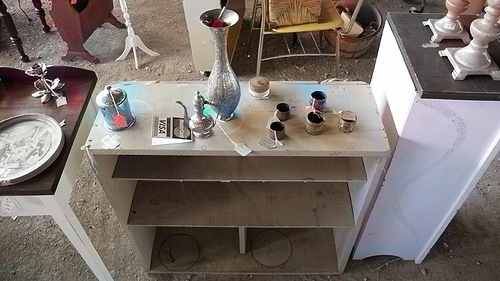}&
\includegraphics[width=0.19\linewidth,height=0.16\linewidth]{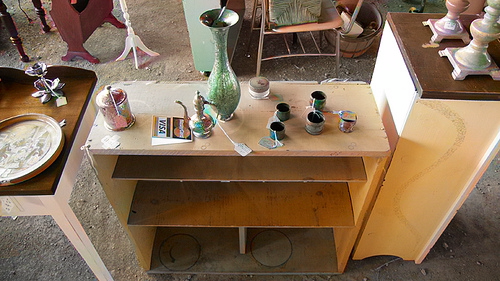}\\

&\includegraphics[width=0.19\linewidth,height=0.16\linewidth]{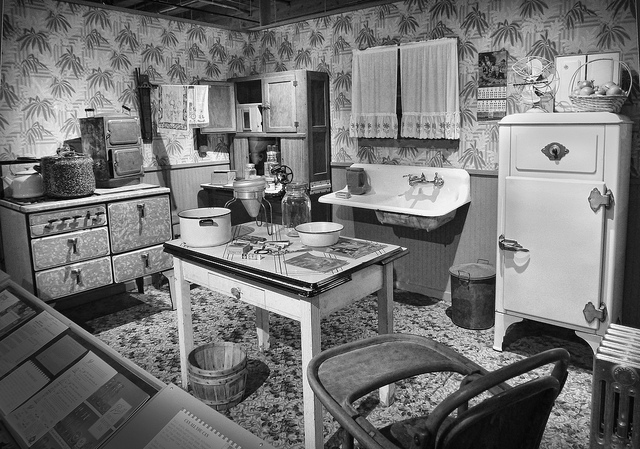}&
\includegraphics[width=0.19\linewidth,height=0.16\linewidth]{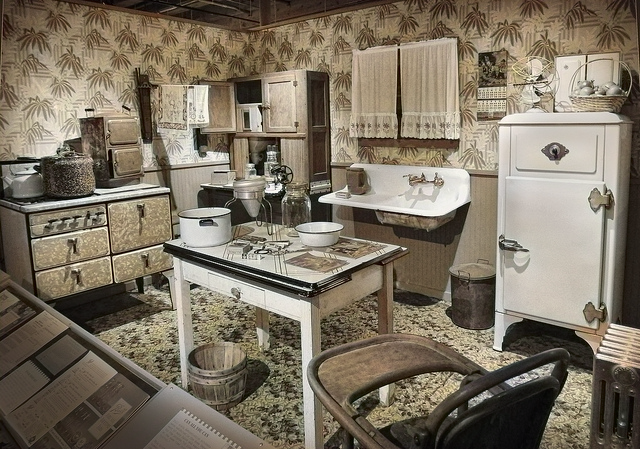}&
\includegraphics[width=0.19\linewidth,height=0.16\linewidth]{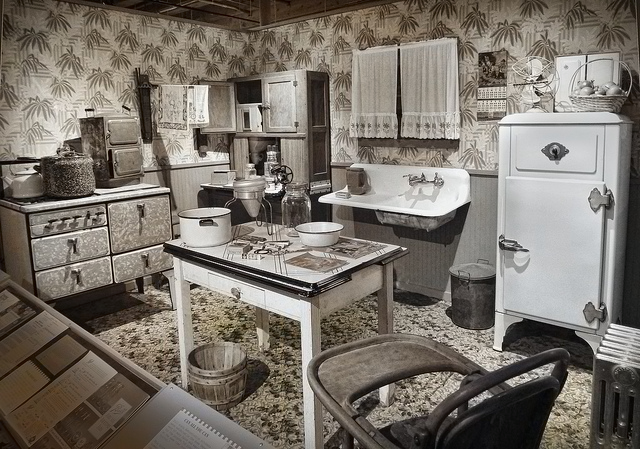}&
\includegraphics[width=0.19\linewidth,height=0.16\linewidth]{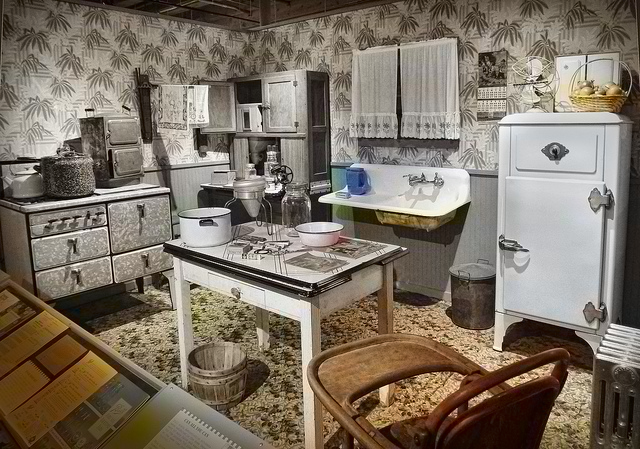}&
\includegraphics[width=0.19\linewidth,height=0.16\linewidth]{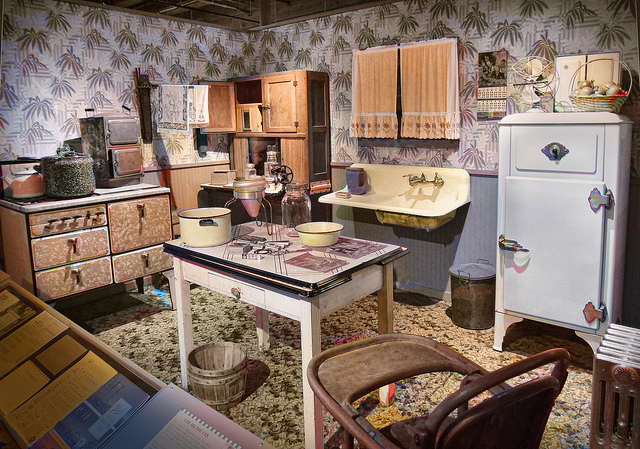}\\

&\includegraphics[width=0.19\linewidth,height=0.16\linewidth]{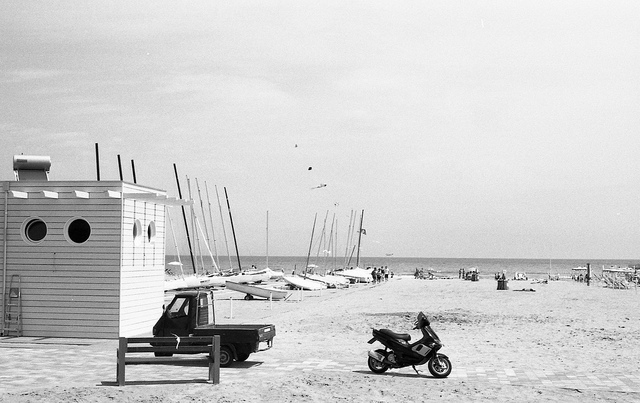}&
\includegraphics[width=0.19\linewidth,height=0.16\linewidth]{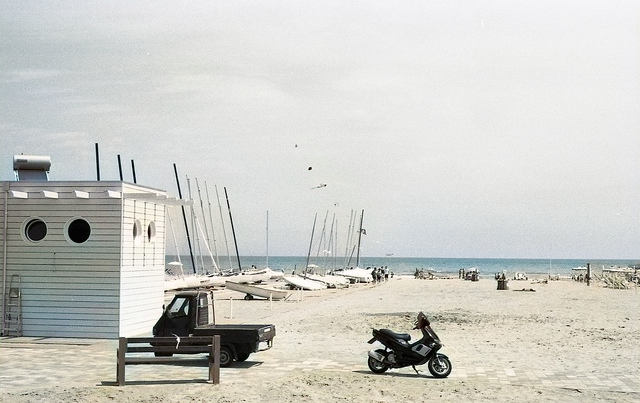}&
\includegraphics[width=0.19\linewidth,height=0.16\linewidth]{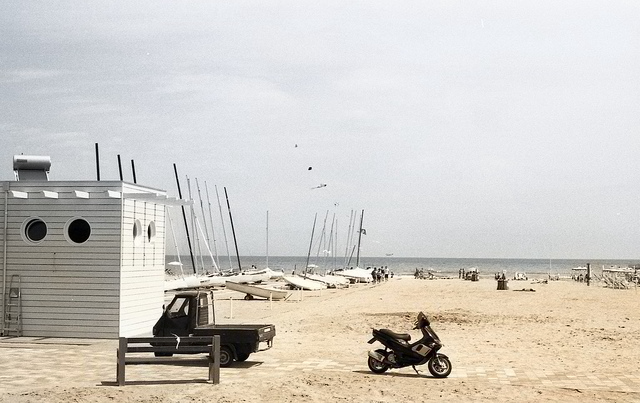}&
\includegraphics[width=0.19\linewidth,height=0.16\linewidth]{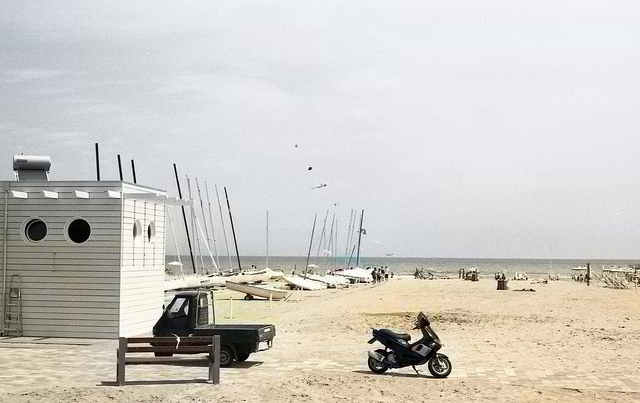}&
\includegraphics[width=0.19\linewidth,height=0.16\linewidth]{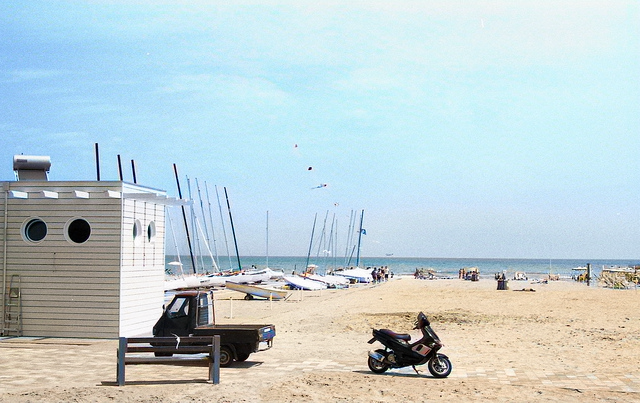}\\

&\includegraphics[width=0.19\linewidth]{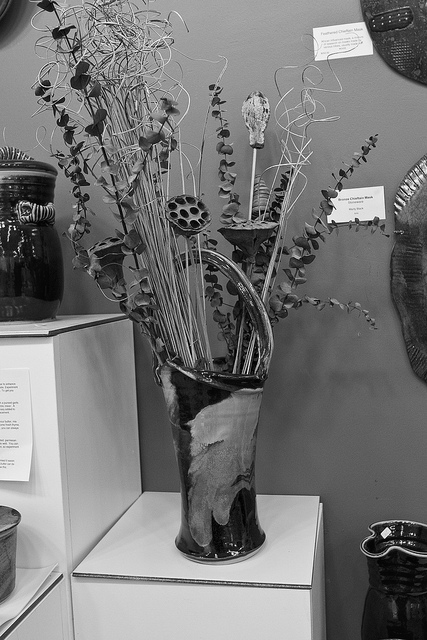}&
\includegraphics[width=0.19\linewidth]{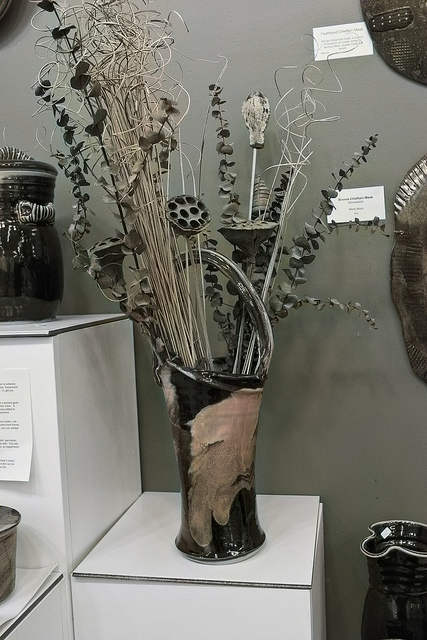}&
\includegraphics[width=0.19\linewidth]{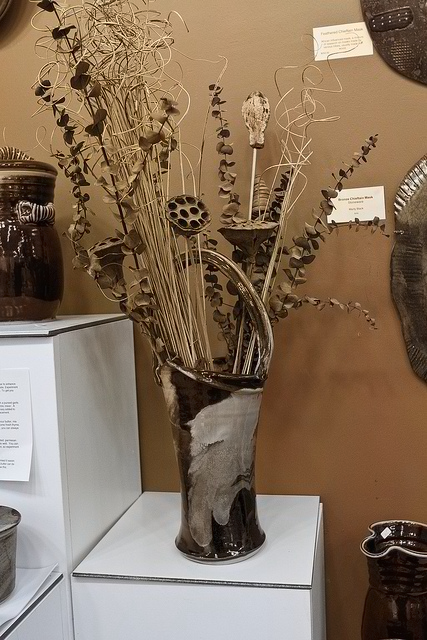}&
\includegraphics[width=0.19\linewidth]{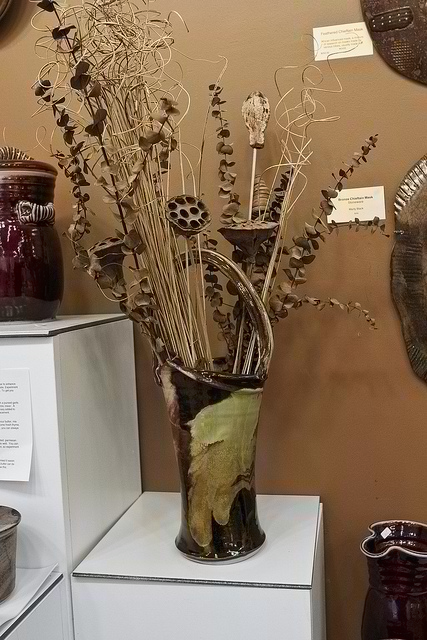}&
\includegraphics[width=0.19\linewidth]{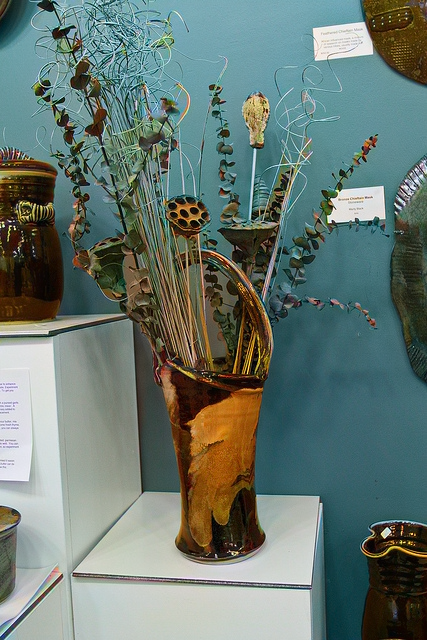}\\
&Input & Deoldify~\cite{deoldify} & Zhang et al.~\cite{Zhang2017} & Su et al.~\cite{su2020instance} & Ours \\
\end{tabular}
\caption{Perceptual comparisons to the state-of-the-art methods.  Our method predicts visually pleasing colors where other methods cannot. More comparisons are presented in the supplementary material.
}
\label{fig:perceptual_comparison}
\end{figure*}

\section{Experiments}
\subsection{Experimental setting}

\textbf{Dataset.} Our experiments are mainly conducted on COCO-stuff dataset~\cite{lin2014coco}. The images in COCO-stuff contain complex objects and scenes. The training set and validation set of COCO-stuff contain 118K and 5K images, respectively. All networks used in our approach are trained on the COCO-stuff dataset. We adopt the original split in our experiments. We also demonstrate qualitative results on ImageNet~\cite{Deng2009}.

% We use two datasets in the evaluations:
% \begin{itemize}
%     \setlength{\itemsep}{0pt}
%     \setlength{\parsep}{0pt}
%     \setlength{\parskip}{0pt}
%     \item[] \textbf{ImageNet~\cite{Deng2009}}:  We use the testing split provided by Larsson et al.~\cite{Larsson2016}, which contains 10,000 images.
%     \item[] \textbf{COCO-stuff}: we use the original validation set.

% \end{itemize}

\textbf{Baselines.} We adopt the most recent and related automatic colorization methods as baselines, including CIC~\cite{Zhang2016}, Zhang et al.~\cite{Zhang2017}, Lei et al.~\cite{Lei_2019_CVPR}, Deoldify~\cite{deoldify}, Su et al.~\cite{su2020instance}. All of them can be used for automatic image colorization. Zhang et al.~\cite{Zhang2017} and Su et al.~\cite{su2020instance} are trained or finetuned on the COCO-stuff dataset. We directly use the released results by Su et al.~\cite{su2020instance} in our experiments.

\subsection{Comparisons to baselines} 
\textbf{Quantitative comparison.} We compute the colorfulness through the metric proposed by Hasler et al.~\cite{hasler2003colorfulness} to compare with existing methods quantitatively. As shown in Table~\ref{table:baseline_comparison}, our approach generates the most colorful results. We do not adopt the PSNR metric since it cannot represent the colorization quality for diverse colorization, as reported by Lei et al.~\cite{Lei_2019_CVPR}. The analysis of PSNR is presented in supplement for reference.

\begin{wraptable}{r}{0.5\linewidth}
\renewcommand{\arraystretch}{1.2}
% \vspace{-2mm}
\centering
\begin{tabular*}{0.5\textwidth}{l@{\hspace{3mm}}c@{\hspace{1mm}}c@{\hspace{1mm}}}
\toprule
& \multicolumn{2}{c}{Preference rate}\\
Comparison & Ours& Baseline \\
\hline
Ours >  Su et al.~\cite{su2020instance} & 76.9\% & 23.1\% \\
Ours > Zhang et al.~\cite{Zhang2017}  & 77.4\% & 22.6\% \\
\bottomrule
\end{tabular*}
\caption{The results of our user study. Our results are preferred by human observers.} 
\label{table:user_study}
% \vspace{-5mm}
\end{wraptable} 
\textbf{User preference comparison.}  We conduct a perceptual experiment by user study to evaluate which approach is preferred by human observers. Specifically, we compare our method with two strong baselines: Zhang et al.~\cite{Zhang2017} and Su et al.~\cite{su2020instance}. A user is presented with a pair of colorized images (our approach and a baseline) simultaneously in each comparison. The positions of our result (i.e., left image or right image) are random in each comparison. We let the user choose the one that is more realistic and colorful. We randomly sample 100 images from the COCO-Stuff validation dataset, and 29 volunteers participated in this user study.

Table~\ref{table:user_study} shows the results of our user study. Our approach is significantly preferred than Zhang et al.~\cite{Zhang2017} and Su et al.~\cite{su2020instance}. Interestingly, we observe that there is no much difference between the preference of Zhang et al.~\cite{Zhang2017} and Su et al.~\cite{su2020instance}. We believe it is because Su et al.~\cite{su2020instance} is based on the model of Zhang et al.~\cite{Zhang2017}. Their qualitative results are also similar, as shown in Figure~\ref{fig:perceptual_comparison}.
%Our user study is the key experiment to evaluate the performance of various methods.

\begin{figure*}[]
\centering
\begin{tabular}{@{}c@{\hspace{1mm}}c@{\hspace{1mm}}c@{\hspace{1mm}}c@{\hspace{1mm}}c@{\hspace{1mm}}c@{}}
%&\includegraphics[width=0.19\linewidth]{figure/diverse/000000023230/gray.png}&
%\includegraphics[width=0.19\linewidth]{figure/diverse/000000023230_lei/000000023230_1.png}&
%\includegraphics[width=0.19\linewidth]{figure/diverse/000000023230_lei/000000023230_2.png}&
%\includegraphics[width=0.19\linewidth]{figure/diverse/000000023230_lei/000000023230_3.png}&
%\includegraphics[width=0.19\linewidth]{figure/diverse/000000023230_lei/000000023230_4.png}\\
%&\includegraphics[width=0.19\linewidth]{figure/diverse/000000023230/result_03.png}&
%\includegraphics[width=0.19\linewidth]{figure/diverse/000000023230/result_00.png}&
%\includegraphics[width=0.19\linewidth]{figure/diverse/000000023230/result_01.png}&
%\includegraphics[width=0.19\linewidth]{figure/diverse/000000023230/result_05.png}&
%\includegraphics[width=0.19\linewidth]{figure/diverse/000000023230/result_08.png}\\
&\includegraphics[width=0.19\linewidth]{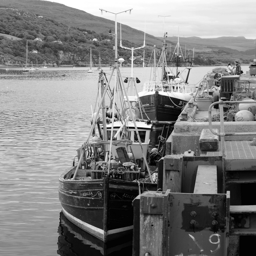}&
\includegraphics[width=0.19\linewidth]{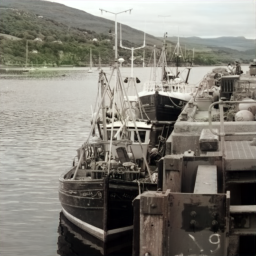}&
\includegraphics[width=0.19\linewidth]{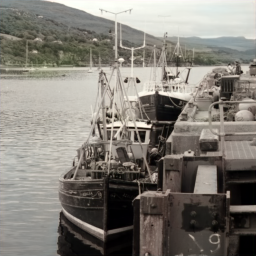}&
\includegraphics[width=0.19\linewidth]{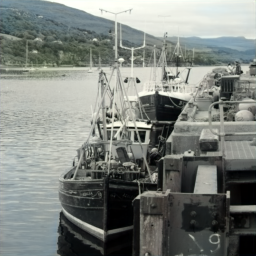}&
\includegraphics[width=0.19\linewidth]{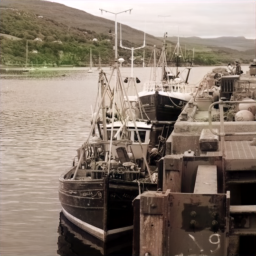}\\
&Input & AVC~\cite{Lei_2019_CVPR}: result 1 & AVC~\cite{Lei_2019_CVPR}: result 2 &AVC~\cite{Lei_2019_CVPR}: result 3 &AVC~\cite{Lei_2019_CVPR}: result 4 \\

% &\includegraphics[width=0.19\linewidth]{figure/diverse/000000020571/synthesis_04.png}&
% \includegraphics[width=0.19\linewidth]{figure/diverse/000000020571/synthesis_00.png}&
% \includegraphics[width=0.19\linewidth]{figure/diverse/000000020571/synthesis_01.png}&
% \includegraphics[width=0.19\linewidth]{figure/diverse/000000020571/synthesis_03.png}&
% \includegraphics[width=0.19\linewidth]{figure/diverse/000000020571/synthesis_08.png}\\
% &Ours: result 1 & Ours: result 2 & Ours: result 3 & Ours: result 4 & Ours: result 5 \\

&\includegraphics[width=0.19\linewidth]{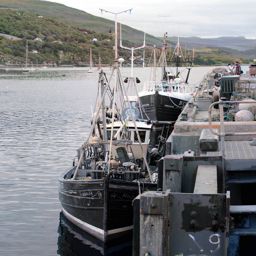}&
\includegraphics[width=0.19\linewidth]{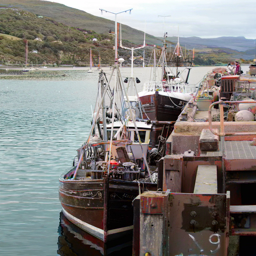}&
\includegraphics[width=0.19\linewidth]{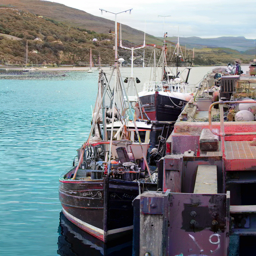}&
\includegraphics[width=0.19\linewidth]{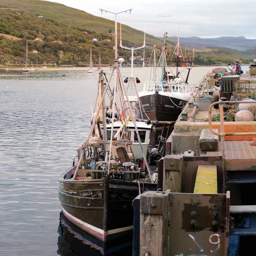}&
\includegraphics[width=0.19\linewidth]{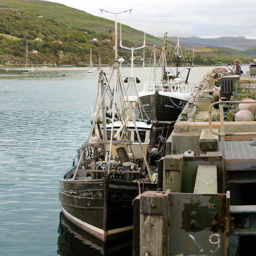}\\
&Ours: result 1 & Ours: result 2 & Ours: result 3 & Ours: result 4 & Ours: result 5 \\
\end{tabular}
\caption{Qualitative comparisons to AVC (Lei et al.~\cite{Lei_2019_CVPR}). AVC is an automatic colorization that considers the diverse possible colorized results. Our results are more diverse and colorful.} %Four diverse results colorized by our approach for the same grayscale image. 
\label{fig:diverse_colorization}
\end{figure*}

\begin{figure}[t]
\centering

\begin{tabular}{@{}c@{\hspace{1mm}}c@{\hspace{1mm}}c@{\hspace{1mm}}c@{\hspace{1mm}}c@{\hspace{1mm}}c@{}}
\includegraphics[width=0.24\linewidth]{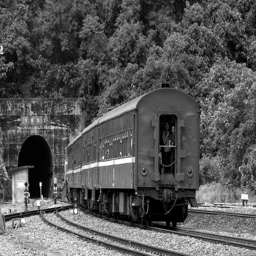}&
\includegraphics[width=0.24\linewidth]{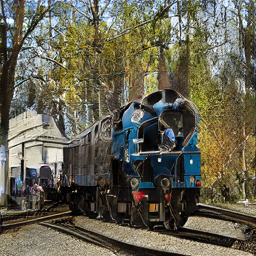}&
\includegraphics[width=0.24\linewidth]{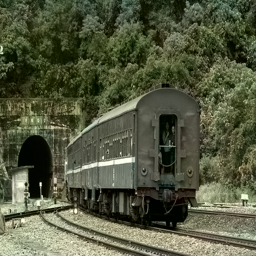}&
\includegraphics[width=0.24\linewidth]{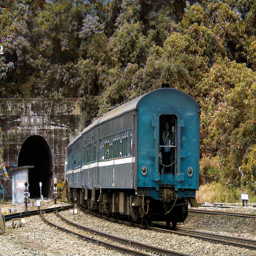}\\
Input & Imagination & Zhang et al.~\cite{zhang2019deepvideocolor} & Ours\\
\end{tabular}
\caption{The qualitative comparison between our model and Zhang et al.~\cite{zhang2019deepvideocolor}. Our colorization network can preserve the original color better.}
\label{fig:Not_a_plus_b}
\end{figure}

\begin{figure}[t]
\centering
\begin{tabular}{@{}c@{\hspace{1mm}}c@{\hspace{1mm}}c@{\hspace{1mm}}c@{\hspace{1mm}}c@{\hspace{1mm}}c@{}}
&\includegraphics[width=0.19\linewidth]{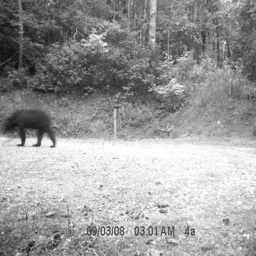}&
\includegraphics[width=0.19\linewidth]{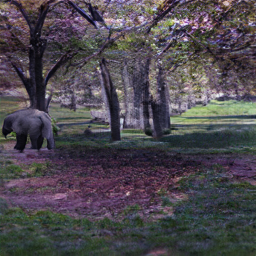}&
\includegraphics[width=0.19\linewidth]{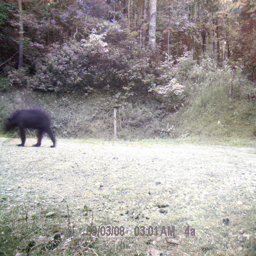}&
\includegraphics[width=0.19\linewidth]{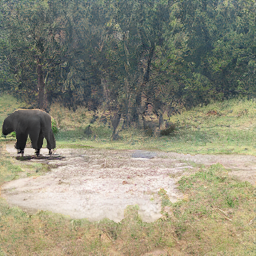}&
\includegraphics[width=0.19\linewidth]{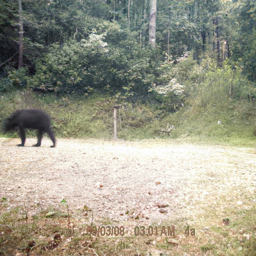}\\
% &\includegraphics[width=0.19\linewidth]{figure/multi_refs/000000565989/gray.png}&
% \includegraphics[width=0.19\linewidth]{figure/multi_refs/000000565989/synthesis_01.png}&
% \includegraphics[width=0.19\linewidth]{figure/multi_refs/000000565989/result_01.png}&
% \includegraphics[width=0.19\linewidth]{figure/multi_refs/000000565989/mix_synthesis.png}&
% \includegraphics[width=0.19\linewidth]{figure/multi_refs/000000565989/mix_result.png}\\
& Input & Single imagination & Result & Multi-imagination & Result\\
\end{tabular}
\caption{Ablation study on multiple references fusion. If we use a single reference as guidance, the synthesized image may mismatch the input, producing undesirable results. Fusion multiple references based on the luminance can significantly improve the quality of reference and produce better results.}
\label{fig:MultipleReference.}
\end{figure}

\textbf{Qualitative comparisons.} Figure~\ref{fig:perceptual_comparison} shows qualitative comparisons among our approach and baselines. Our results are more colorful than baselines.

\textbf{Diverse colorization.} There are some methods that enable diverse colorization such as Lei et al.~\cite{Lei_2019_CVPR}, Deshande et al.~\cite{Deshpande2017diverse}. However, as shown in Table~\ref{table:baseline_comparison}, most automatic colorization methods cannot generate diverse results. Here we compare our approach with Lei et al.~\cite{Lei_2019_CVPR} in Figure~\ref{fig:diverse_colorization}. While the results of Lei et al.~\cite{Lei_2019_CVPR} are different, each result is not that colorful. As a comparison, our results are more colorful and diverse.

\begin{figure}[]
\centering

\begin{tabular}{@{}c@{\hspace{1mm}}c@{\hspace{1mm}}c@{\hspace{1mm}}c@{\hspace{1mm}}c@{\hspace{1mm}}c@{}}
\includegraphics[width=0.24\linewidth]{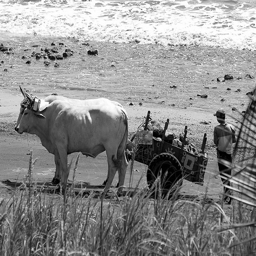}&
\includegraphics[width=0.24\linewidth]{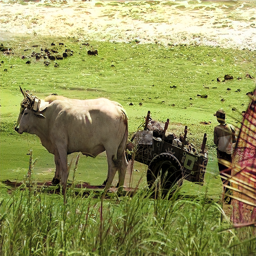}&
\includegraphics[width=0.24\linewidth]{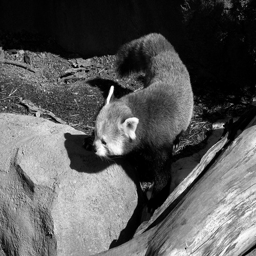}&
\includegraphics[width=0.24\linewidth]{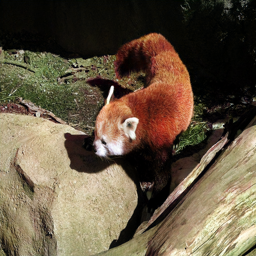}\\
Input & Result & Input & Result \\
\end{tabular}
\caption{Qualitative results on the ImageNet dataset~\cite{Deng2009}. The reference images are generated by another imagination module introduced in the main text. }
\label{fig:ImageNet}
\end{figure}

\subsection{Analysis}
\label{subsec:Controlled experiment}
\textbf{Comparison to reference-based colorization.} While our imagination module can be combined with an existing reference-based image colorization method~\cite{He2018},  we notice existing reference-based methods tend to reduce the colorfulness or change the color of reference image. Figure~\ref{fig:Not_a_plus_b} shows an example for this phenomenon. As mentioned in Section 5, our designed colorization module can preserve the original color better. %In addition, they can also receive one single reference image as input: as mentioned before, it is almost impossible to find a reference image contain all the things for a complex scene. 

% In addition, they can also receive one single reference image as input: as mentioned before, it is almost impossible to find a reference image contain all the things for a complex scene. %We simply combine off-the-shelf models as a baseline. Specifically%

\textbf{Multiple references.} We implement another version by replacing the multiple references with a single synthesized image for comparison. Figure~\ref{fig:MultipleReference.} shows examples of qualitative comparison. If we only use a single reference image, sometimes the color of the synthesized image does not match the input, leading to unsatisfying colorization results. Using multiple reference images to compose a new image improves the robustness of our approach.
%We analyze the importance of using multiple references here. 

\textbf{Other imagination modules.} As discussed in Section 4, the representation of context in an image can be quite diverse. We try another type of imagination module on ImageNet~\cite{Deng2009}. Specifically, the context extraction network $c$ is an image detection network, which helps us obtain the class and location. Secondly, given a label of class, we use BigGAN~\cite{biggan} as the synthesis model $g$ to synthesize images. We then compose the objects to a new reference image $\mathbf{R}$ through the locations. The qualitative results are presented in Figure~\ref{fig:ImageNet}. Details of this implementation are presented in the supplementary material.

\textbf{Limitations.} The colorized results heavily depend on the images from our imagination modules. When the imaginations are bad, our colorized results can also be bad. More analysis is presented in the supplementary material due to limited space.
% Our approach is robust in most cases. But we notice a 
% Our imagination module needs to extract the semantic information first, and this process can have mistakes (obviously, the correctness of image classification or semantic segmentation is not one hundred percent). However, we find an interesting phenomenon that when our algorithm fails, other baselines might also fail. 

\subsection{Colorizing legacy photographs}
We present qualitative results for colorizing legacy photographs in Figure~\ref{fig:legacy}. While these photos were captured tens of years or one hundred years ago, our approach can still generate photorealistic colorized results. More results are presented in the supplementary material.

\section{Discussion}
We have presented a novel approach to automatic image colorization by imitating the imagination process of human experts. We propose an imagination module that can synthesize diverse and colorful reference images with similar contexts of input images. Then a dedicated colorization module is designed to colorize the black-and-white images with the guidance of imagination. Our results can be diverse and realistic with the imagination module. We believe that the concept of imagination can also be applied to various computer vision tasks.

{
\small
\bibliographystyle{plain}
\bibliography{neurips_2021}
}

% [1] Alexander, J.A.\ \& Mozer, M.C.\ (1995) Template-based algorithms for
% connectionist rule extraction. In G.\ Tesauro, D.S.\ Touretzky and T.K.\ Leen
% (eds.), {\it Advances in Neural Information Processing Systems 7},
% pp.\ 609--616. Cambridge, MA: MIT Press.

% [2] Bower, J.M.\ \& Beeman, D.\ (1995) {\it The Book of GENESIS: Exploring
%   Realistic Neural Models with the GEneral NEural SImulation System.}  New York:
% TELOS/Springer--Verlag.

% [3] Hasselmo, M.E., Schnell, E.\ \& Barkai, E.\ (1995) Dynamics of learning and
% recall at excitatory recurrent synapses and cholinergic modulation in rat
% hippocampal region CA3. {\it Journal of Neuroscience} {\bf 15}(7):5249-5262.

%%%%%%%%%%%%%%%%%%%%%%%%%%%%%%%%%%%%%%%%%%%%%%%%%%%%%%%%%%%%
\section*{Checklist}

% %%% BEGIN INSTRUCTIONS %%%
% The checklist follows the references.  Please
% read the checklist guidelines carefully for information on how to answer these
% questions.  For each question, change the default \answerTODO{} to \answerYes{},
% \answerNo{}, or \answerNA{}.  You are strongly encouraged to include a {\bf
% justification to your answer}, either by referencing the appropriate section of
% your paper or providing a brief inline description.  For example:
% \begin{itemize}
%   \item Did you include the license to the code and datasets? \answerYes{See Section~\ref{gen_inst}.}
%   \item Did you include the license to the code and datasets? \answerNo{The code and the data are proprietary.}
%   \item Did you include the license to the code and datasets? \answerNA{}
% \end{itemize}
% Please do not modify the questions and only use the provided macros for your
% answers.  Note that the Checklist section does not count towards the page
% limit.  In your paper, please delete this instructions block and only keep the
% Checklist section heading above along with the questions/answers below.
% %%% END INSTRUCTIONS %%%

\begin{enumerate}

\item For all authors...
\begin{enumerate}
   \item Do the main claims made in the abstract and introduction accurately reflect the paper's contributions and scope?
     \answerYes{}
   \item Did you describe the limitations of your work?
     \answerYes{}
   \item Did you discuss any potential negative societal impacts of your work?
     \answerNo{}
   \item Have you read the ethics review guidelines and ensured that your paper conforms to them?
     \answerYes{}
 \end{enumerate}

 \item If you are including theoretical results...
 \begin{enumerate}
   \item Did you state the full set of assumptions of all theoretical results?
     \answerYes{}
 	\item Did you include complete proofs of all theoretical results?
     \answerYes{}
 \end{enumerate}

 \item If you ran experiments...
 \begin{enumerate}
   \item Did you include the code, data, and instructions needed to reproduce the main experimental results (either in the supplemental material or as a URL)?
     \answerNo{}
   \item Did you specify all the training details (e.g., data splits, hyperparameters, how they were chosen)?
     \answerYes{}
 	\item Did you report error bars (e.g., with respect to the random seed after running experiments multiple times)?
     \answerNo{}
 	\item Did you include the total amount of compute and the type of resources used (e.g., type of GPUs, internal cluster, or cloud provider)?
     \answerYes{}
 \end{enumerate}

 \item If you are using existing assets (e.g., code, data, models) or curating/releasing new assets...
 \begin{enumerate}
   \item If your work uses existing assets, did you cite the creators?
     \answerYes{}
   \item Did you mention the license of the assets?
     \answerNA{}
   \item Did you include any new assets either in the supplemental material or as a URL?
     \answerNo{}
   \item Did you discuss whether and how consent was obtained from people whose data you're using/curating?
     \answerNA{}
   \item Did you discuss whether the data you are using/curating contains personally identifiable information or offensive content?
     \answerNA{}
 \end{enumerate}

 \item If you used crowdsourcing or conducted research with human subjects...
 \begin{enumerate}
   \item Did you include the full text of instructions given to participants and screenshots, if applicable?
     \answerYes{}
   \item Did you describe any potential participant risks, with links to Institutional Review Board (IRB) approvals, if applicable?
     \answerNA{}
   \item Did you include the estimated hourly wage paid to participants and the total amount spent on participant compensation?
     \answerYes{}
 \end{enumerate}

 \end{enumerate}

% %%%%%%%%%%%%%%%%%%%%%%%%%%%%%%%%%%%%%%%%%%%%%%%%%%%%%%%%%%%%

% \appendix

% \section{Appendix}

% Optionally include extra information (complete proofs, additional experiments and plots) in the appendix.
% This section will often be part of the supplemental material.

\end{document}